\documentclass[conference]{IEEEtran}
\usepackage{times}

\usepackage[numbers]{natbib}
\usepackage{multicol}
\usepackage[bookmarks=true]{hyperref}
\usepackage{xcolor}
\usepackage{graphicx} 
\usepackage{amsmath}
\usepackage{adjustbox}
\usepackage{subcaption}
\usepackage{amssymb}
\usepackage{xspace}

\begin{document}
\newcommand{\todo}[1]{\textcolor{red}{#1}}

\title{Self-Supervised Goal-Conditioned Pick and Place}

\author{\authorblockN{Coline Devin}
\authorblockA{ UC Berkeley\\
coline@eecs.berkeley.edu}
\and
\authorblockN{Payam Rowghanian}
\authorblockA{Osaro Inc.\\
payam@osaro.com}
\and
\authorblockN{Chris Vigorito}
\authorblockA{Osaro Inc.\\
chris@osaro.com}
\and
\authorblockN{Will Richards}
\authorblockA{Osaro Inc.\\
will@osaro.com}
\and
\authorblockN{Khashayar Rohanimanesh}
\authorblockA{Osaro Inc.\\
khash@osaro.com}
}

\maketitle
\newcommand{\impick}{{I_{\text{grasp}}}}
\newcommand{\implace}{{I_{\text{place}}}}
\newcommand{\imgripper}{{I_{\text{wrist}}}}
\newcommand{\gp}{{(x, y)_g}}
\newcommand{\pp}{{(x, y)_p}}
\newcommand{\ee}{{(x, y)_{w}}} %
\newcommand{\encbin}{\Phi_{B}}
\newcommand{\encee}{\Phi_{W}}

\newcommand{\imgoal}{{I_{\text{goal}}}}
\newcommand{\impart}{{I_{\text{parts}}}}
\newcommand{\imkit}{{I_{\text{kit}}}}

\newcommand{\flatlay}{\textbf{flatlay}\xspace}
\newcommand{\cluttered}{\textbf{cluttered}\xspace}

\newcommand{\fullimage}{\textbf{Full Image}\xspace}
\newcommand{\gam}{\textbf{Gamma}\xspace}
\newcommand{\grasplace}{\textbf{Grasp:Place}\xspace}

\newcommand{\embee}{\phi_\text{w}}
\newcommand{\embpart}[1]{\phi^{#1}_\text{grasp}}
\newcommand{\embkit}[1]{\phi^{#1}_\text{place}}
\newcommand{\embplace}[1]{\phi^{#1}_\text{place}}
\newcommand{\embgrasp}[1]{\phi^{#1}_\text{grasp}}

\newcommand{\embgoal}[1]{\phi^{#1}_\text{goal}}

\begin{abstract}
Robots have the capability to collect large amounts of data autonomously by interacting with objects in the world. However, it is often not obvious \emph{how} to learning from autonomously collected data without human-labeled supervision. In this work we learn pixel-wise object representations from unsupervised pick and place data that generalize to new objects. We introduce a novel framework for using these representations in order to predict where to pick and where to place in order to match a goal image. Finally, we demonstrate the utility of our approach in a simulated grasping environment.
\end{abstract}

\IEEEpeerreviewmaketitle

\section{Introduction}
Industrial robotics uses, such as piece-picking and kitting, need robots to be able to manipulate diverse objects. Human-supervised learning based-approaches for detecting objects have shown promise in robotics.
However, given the immense variety of objects in the world, requiring  human-annotated segmentation data or CAD models for all objects a robot may face is not a scalable approach. Learning visual representations of objects \emph{without} human annotations could be cheaper and provide faster adaptability to novel objects. Robots offer promising avenues for self supervised learning as they interact with the world and record data. By taking actions and changing the state of their environment, robots generate supervision signals without relying on human annotators.

Self-supervised learning has been approached from various directions in the context of robotics. Time-contrastive methods use multiple viewpoints or simulations of a trajectory to learn features over the progress of task~\cite{sermanet2018time,maeda2020visual,jeong2019self,goodrich2020depth}. Another approach is to move objects around in order to learn about their visual appearance~\cite{florence2018dense,pathakCVPRW18segByInt,sundaresan2020learning}.

We use contrastive learning to learn object-embeddings from unsupervised grasping data in order to perform goal-conditioned grasping and placing. Several prior approaches have shown that robotic grasping data can be used to learn object embeddings useful for goal-conditioned grasping~\cite{jang2018grasp2vec} and assembly~\cite{zakka2019form2fit}. 
These approaches learn useful high dimensional embeddings of object images by moving objects with a robot. For example, by grasping an object the robot has generated two views of an object: the view of the object in the workspace, and the view in the gripper. Contrastive learning~\cite{hadsell2006dimensionality} can then be used to push the embeddings of these two views closer together and the embeddings of other points farther away. We differ from prior work by learning a unified embedding space to solve both grasping and placing simultaneously without relying on assembly priors. One appealing aspect of this framework is that it does not need an additional data collection step specific to this problem beyond normal pick and place datasets which are widely generated as more robots are employed in standard pick and place scenarios across many warehouse environments.

\begin{figure}
    \centering
\begin{subfigure}[t]{0.47\linewidth}
    \centering
    \includegraphics[width=\linewidth]{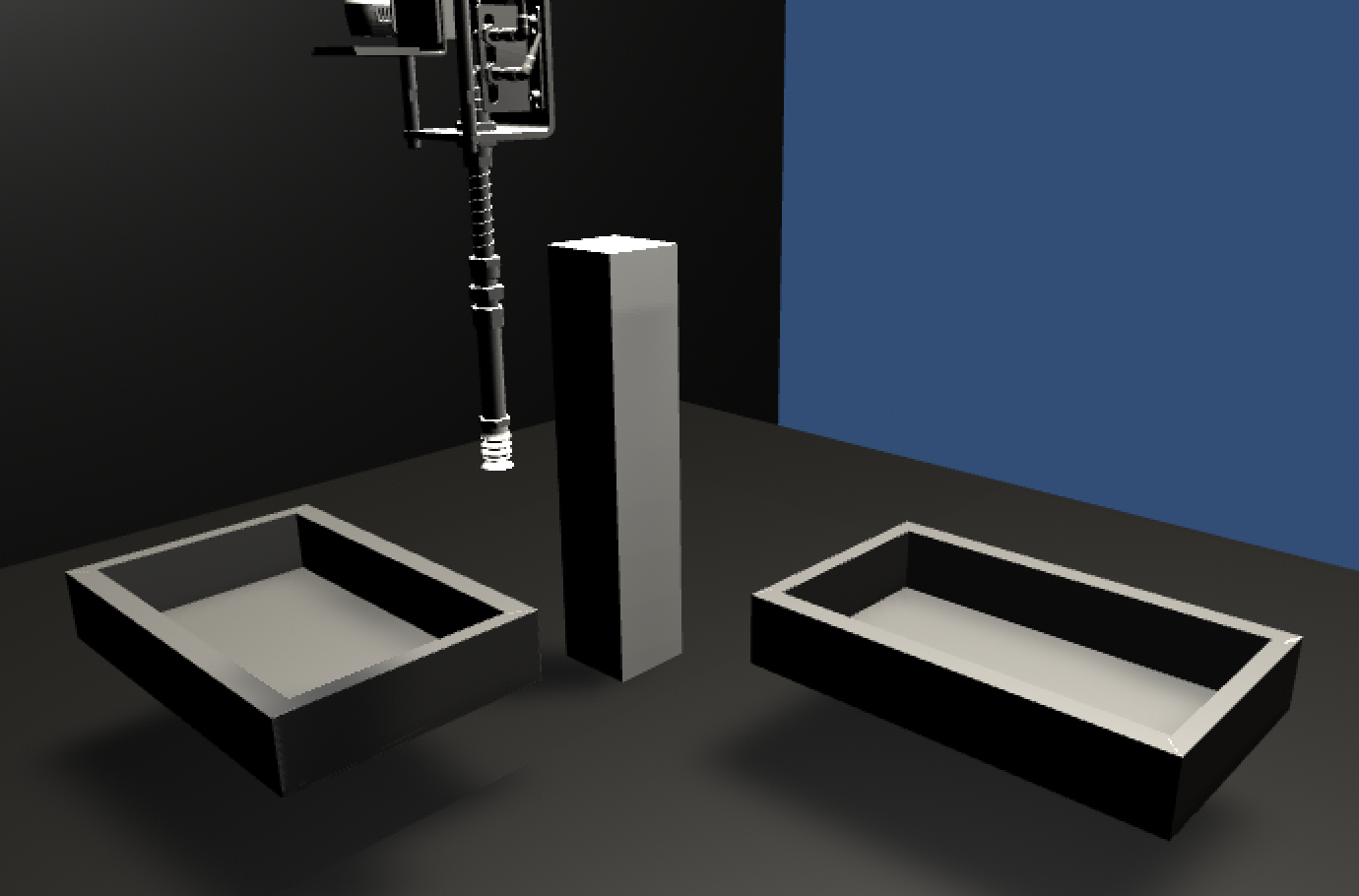}
    \caption{Robot cell}\label{fig:setup}
  \end{subfigure}
    \begin{subfigure}[t]{0.47\linewidth}
    \centering
    \includegraphics[width=0.66\linewidth]{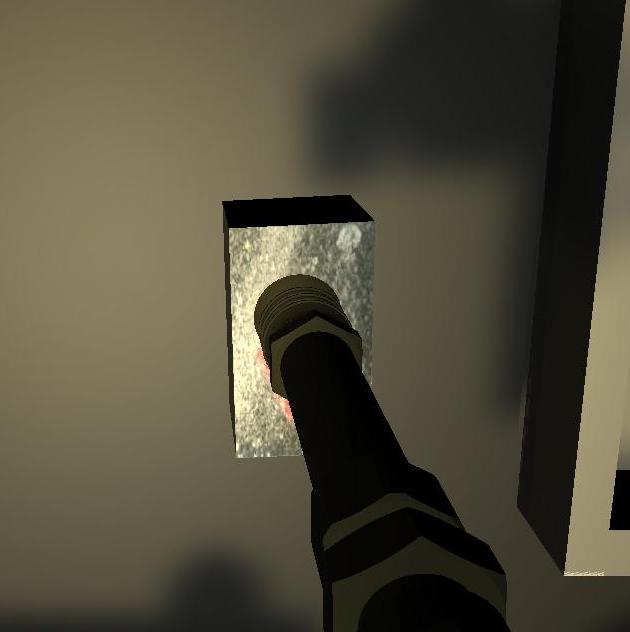}
    \caption{Gripper image, $\ee$ is the center pixel. $\imgripper$}\label{fig:gripper}
  \end{subfigure}\\
\begin{subfigure}[t]{0.47\linewidth}
    \centering
    \includegraphics[width=0.8\linewidth]{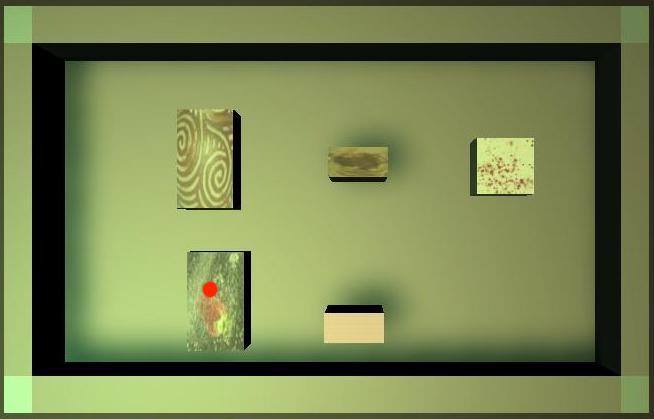}
    \caption{Grasp bin $\impick$ with grasp point $\gp$ shown in red.}\label{fig:pickenv}
  \end{subfigure}
  \begin{subfigure}[t]{0.47\linewidth}
    \centering
    \includegraphics[width=0.8\linewidth]{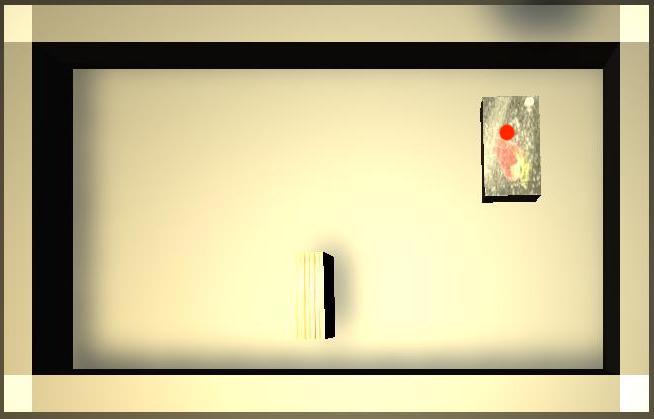}
    \caption{Place bin $\implace$ with place point $\pp$ shown in red.}\label{fig:placeenv}
  \end{subfigure}
    \caption{Robot cell for grasping and placing objects.}
    \label{fig:env}
\end{figure}
Our contributions can be summarized as follows: (1) we present an end-to-end self-supervised framework for learning grasp embeddings that predict grasp and place poses conditioned on a goal image; (2) we present results in a simulated domain and show that our model generalizes to unseen objects.

\begin{figure*}[]
    \centering
    \adjustbox{max width=0.95\linewidth}{
    \begin{tabular}{ccccc}
     Place Bin Before & Goal Image & Grasp Bin & Wrist Cam & Place Bin After  \\
     \includegraphics[width=0.2\linewidth]{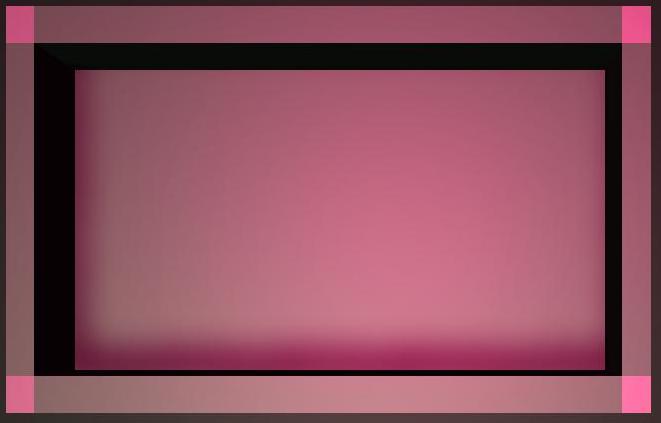}&
       \includegraphics[trim=35 32 23 23, clip,width=0.2\linewidth]{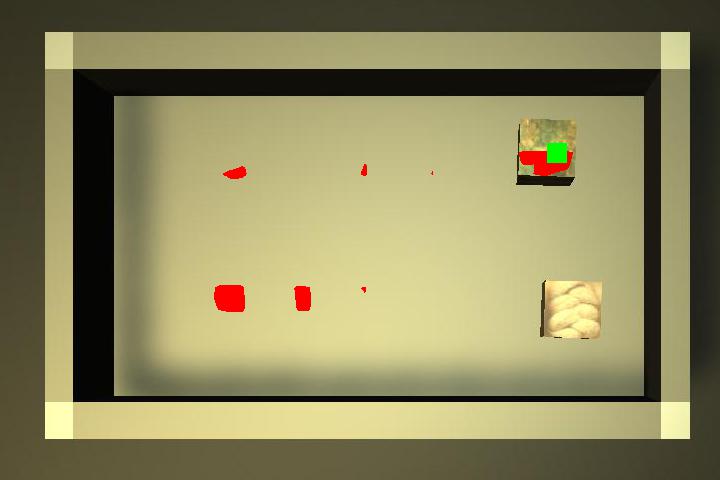}&\includegraphics[trim=32 27 27 30, clip,width=0.2\linewidth]{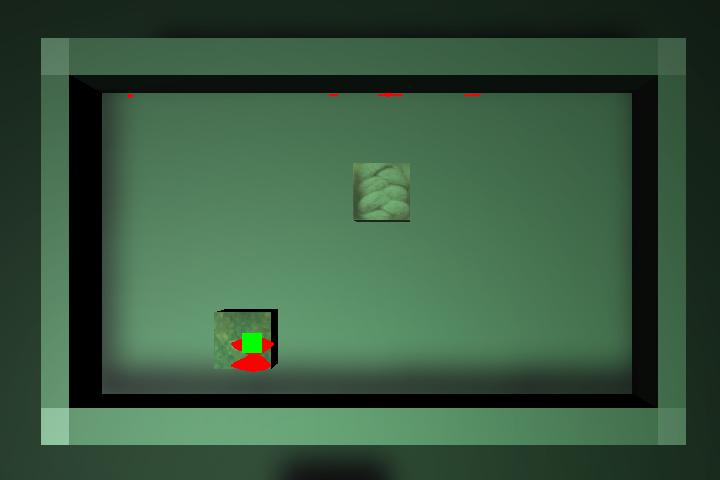}  & \includegraphics[width=0.128\linewidth]{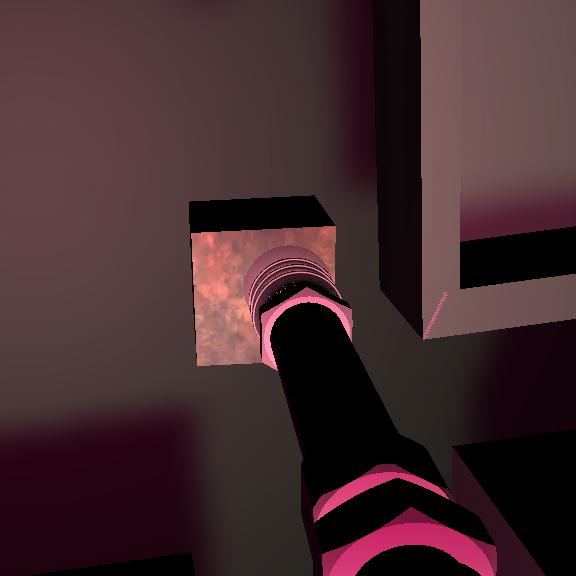} &
    \includegraphics[trim=35 32 23 23, clip,width=0.2\linewidth]{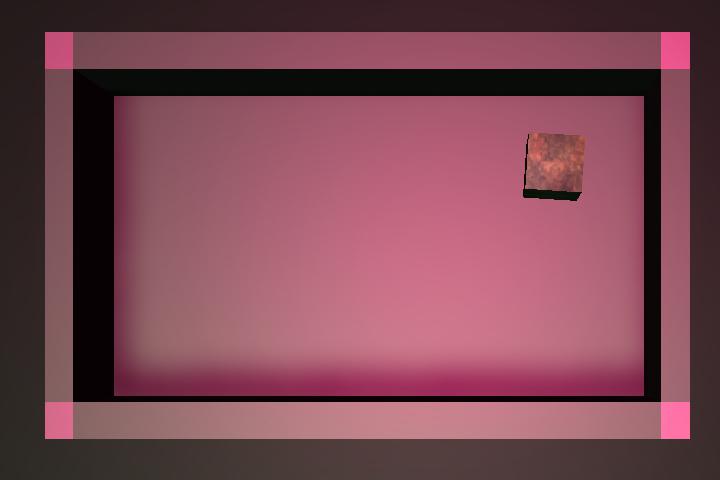} \\
      \includegraphics[width=0.2\linewidth]{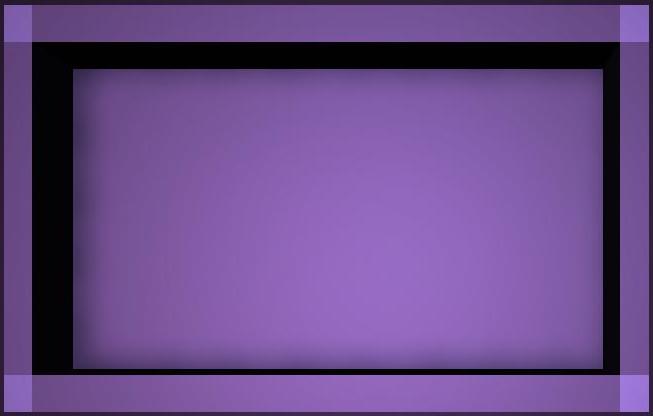}&
  \includegraphics[trim=32 27 27 30, clip,width=0.2\linewidth]{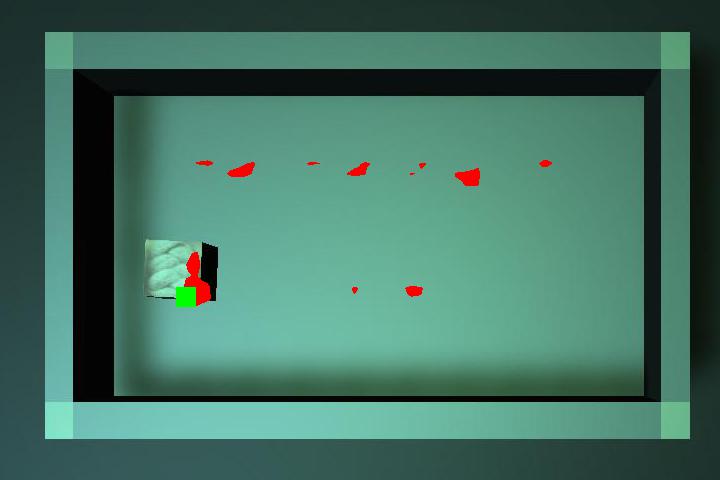}&
\includegraphics[trim=35 32 23 23, clip,width=0.2\linewidth]{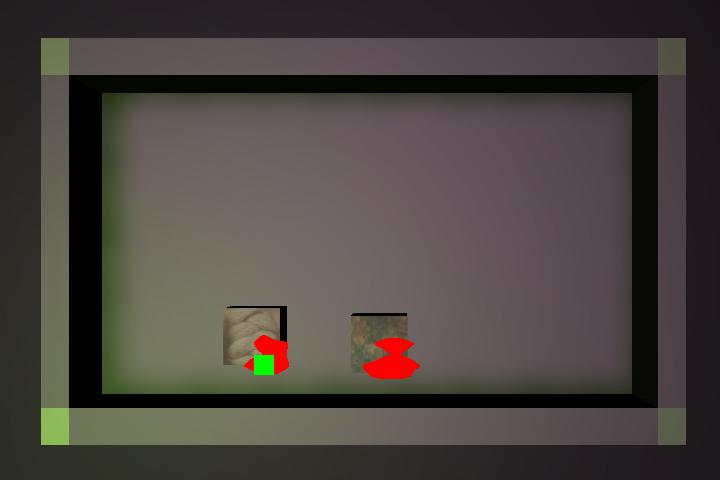}&
\includegraphics[width=0.128\linewidth]{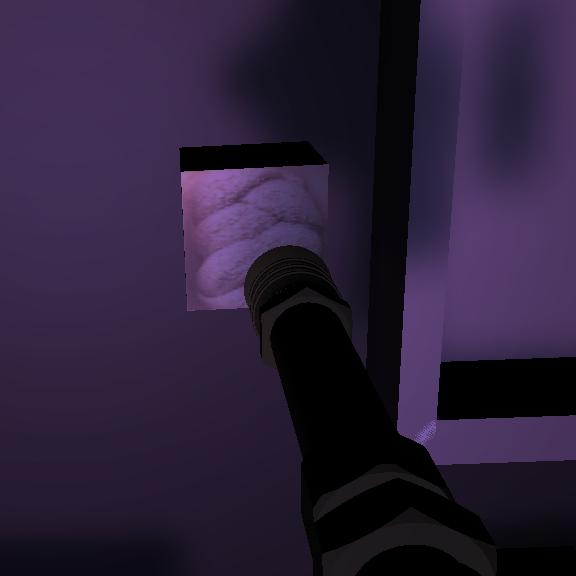}&
\includegraphics[trim=35 32 23 23, clip,width=0.2\linewidth]{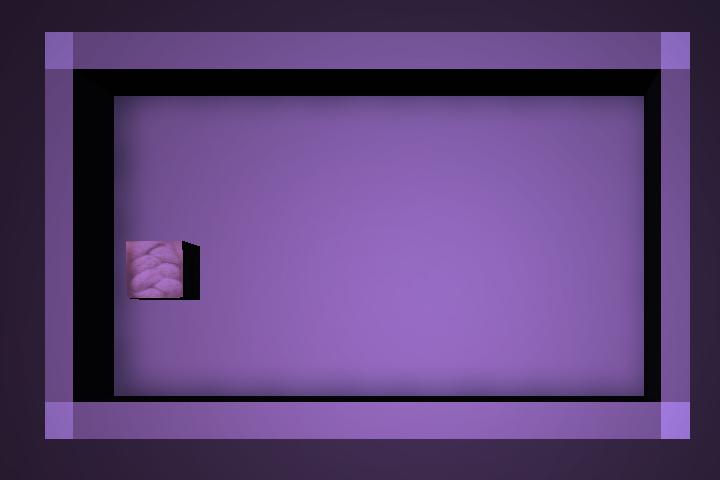}  \\
\includegraphics[width=0.2\linewidth]{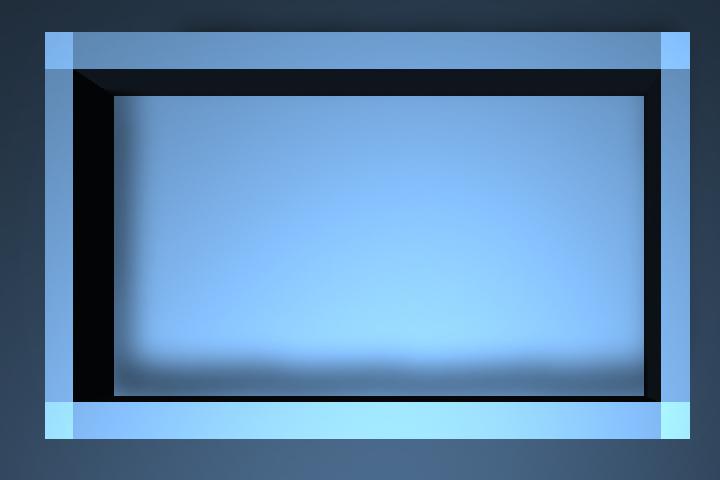}&
  \includegraphics[trim=32 27 27 30, clip,width=0.2\linewidth]{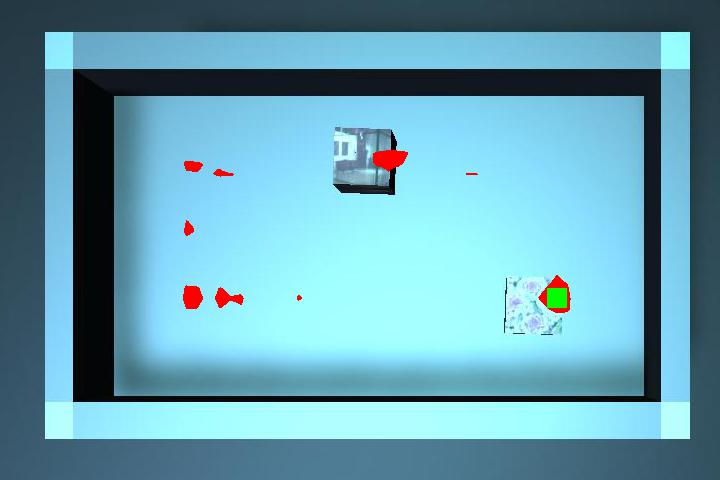}&
\includegraphics[trim=35 32 23 23, clip,width=0.2\linewidth]{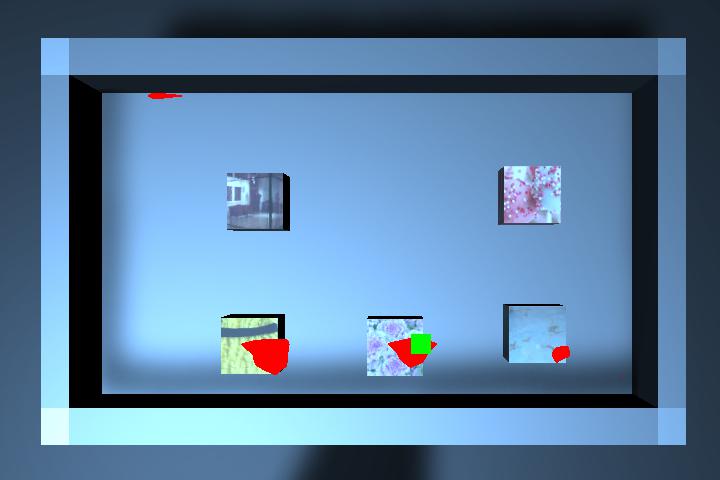}&
\includegraphics[width=0.128\linewidth]{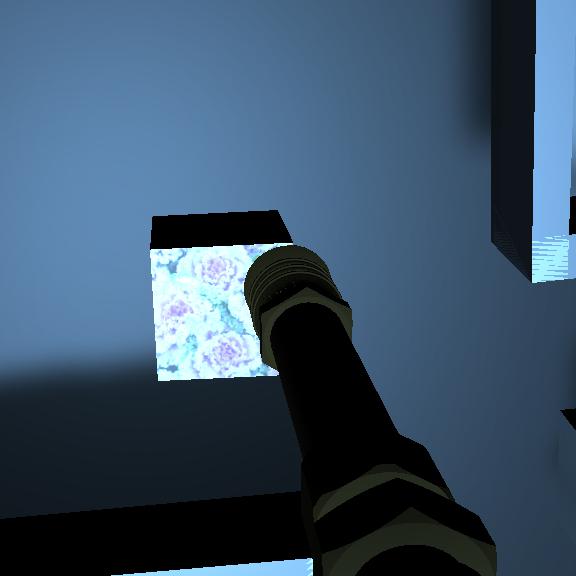}&
\includegraphics[trim=35 32 23 23, clip,width=0.2\linewidth]{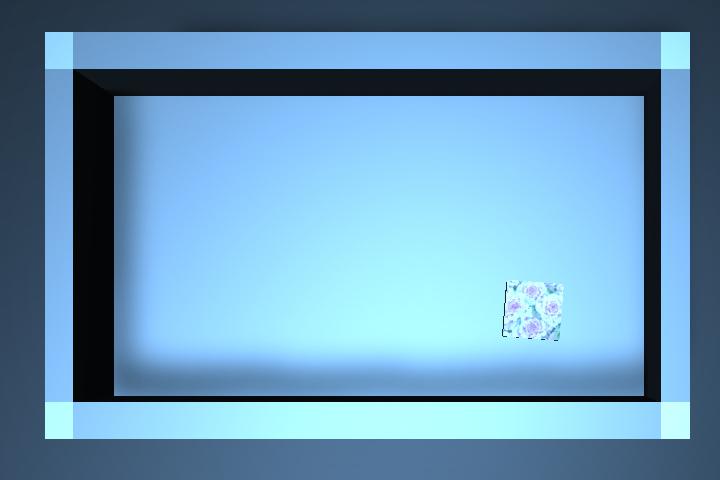}

    \end{tabular}}
    \caption{Examples of goal-conditioned pick and place. The red areas mark the top 10\% of pixels that maximize Equation~\ref{eq:pick} (the green squares show the argmax) with our learned model. We grasp at the green square in the grasp bin, and then use Equation~\ref{eq:place} to choose where to place. The place bin after placing is shown on the right. These examples show how our embeddings learned from random grasps can be used. The color of the lights is randomized for each trial.}
    \label{fig:kitting}
\end{figure*}

\section{Learning pixel-wise object embeddings without supervision}
\begin{figure}[t]
    \centering
    \includegraphics[width=\linewidth]{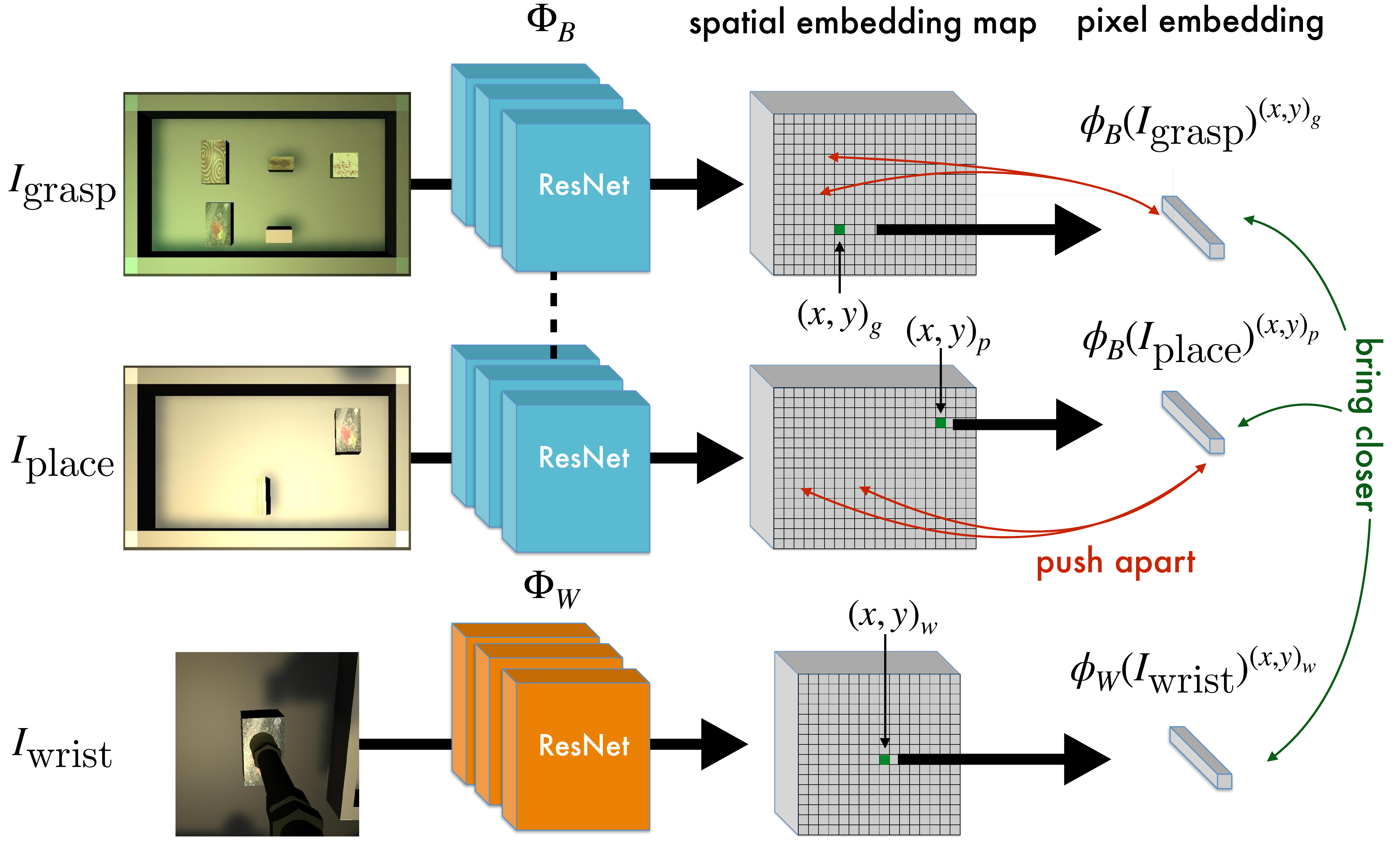}
    \caption{Our model encodes images into pixelwise embeddings. The grasp and place positions provide supervision about which pixels should have similar embeddings across images. }
    \label{fig:model}
\end{figure}

We focus on grasping with suction cups, which provides pixel-level supervision about where an object was grasped and placed. We aim to learn embeddings from grasping data that can be used for deciding both where to grasp and where to place given an image of a goal.

 As illustrated in Figure~\ref{fig:env}, our environment consists of a robotic arm, a source bin, and a target bin (Figure~\ref{fig:setup}). There are three cameras mounted: fixed overhead cameras for each bin (Figures~\ref{fig:pickenv} and~\ref{fig:placeenv}), and also a wrist-cam mounted on the end-effector that captures an image of the grasped object (Figures~\ref{fig:gripper}). An episode of the manipulation task consists of picking an object from the grasp bin conditioned on the placement goal and placing it in the place bin.
In this environment, we collect a dataset of an image of the grasp bin before grasping $\impick$, an image of the place bin after place  $\implace$, and a wrist-cam image of the grasped object in the gripper $\imgripper$. We also record the grasp $\gp$ and the place $\pp$ pixel positions, as well as the pixel position of the end-effector $\ee$ in the wrist image. For performing goal-conditioned actions, we will also have a goal image $\imgoal$ which show a goal configuration of objects in the place bin.

We aim to learn an image encoder $\phi$ which embeds images into a spatial feature map that is useful for goal conditioned grasping and placing. Given a bin of objects and novel goal object configuration, this embedding should indicate where to grasp from the bin and where to place in the target area. In order to learn this from random pick and place data, we use a contrastive representation learning approach, as illustrated in Figure~\ref{fig:model}. We initialize 2 encoders, using the ResNet35 architecture, $\encbin$ and $\encee$ which are applied to bin images ($\impick$ and $\implace$) and the wrist image $\imgripper$, respectively. These encoders are fully convolutional. They share the same architecture, but have separate weights. The encoders reduce the size of the images by a factor of 8 (512 to 64 on the long edge). Operations over pixel embeddings are done at this reduced dimension. 

\subsection{Contrastive Metric Learning}
\newcommand{\graspemb}{\embgrasp{{\gp}}}
\newcommand{\placeemb}{\embplace{{\pp}}}
\newcommand{\wristemb}{\embee}
\newcommand{\negemb}[1]{v_{n_{#1}}}
Our learning objective for the encoders is that the grasp pixel, place pixel, and end-effector pixel should have similar embeddings while being different from the embeddings at other pixels in the images. Similarity is measured as the dot product of the two embedding vectors, which allows the model to output vectors of small magnitude for regions which are never grasped, such as the background. To optimize for this we use a contrastive approach with a cross entropy loss. We simplify notation and use the superscript $(x,y)$ to index into the spatial embedding map output by $\Phi$.
\begin{align*}
     \embgrasp{x,y}=&\encbin(\impick)^{(x,y)}\\
     \embplace{x,y}=& \encbin(\implace)^{(x,y)}\\
     \embee =& \encee(\imgripper)^{\ee}
\end{align*}

Then, $\graspemb$ is the embedding at the grasp pixel in the pick bin image,  $\placeemb$ is the embedding at the place pixel in the place bin image, and $\wristemb$ is the embedding at the end-effector in the wrist image. We want to maximize ${\graspemb}^\top\wristemb$, ${\placeemb}^\top\wristemb$, and ${\graspemb}^\top \placeemb$ while minimizing the dot product of these vectors with embeddings at other pixels. To do this, we sample pixels from the bin images as ``negatives" for the contrastive loss. Let $\negemb{g,k}$ for $k \in [1,K]$ be the embeddings of $K$ ``negative" pixels in the pick bin image. The contrastive loss is implemented as a classification using negative log likelihood. For an anchor embedding $a$, a positive $p$ and negatives $n_{1:K}$
\[\mathcal{L}_\text{con}(a, p, n) = -\log\left(\frac{e^{{a}^\top p}}{e^{{a}^\top p} + \sum_{k=1}^K e^{{a}^\top n_k}}\right).
\]
We also apply a regularization loss to the magnitudes of the embeddings. The regularization is only applied to embeddings with magnitude greater than 1, and is 
\[ \mathcal{L}_{\text{reg}}(v) =   \begin{cases}
                                   ||v||_2 & \text{if $||v||_2>1$} \\
                                   0 & \text{otherwise}
  \end{cases}
\]

The total loss over all the images is then,
\[
\begin{split}
\mathcal{L} = &\mathcal{L}_\text{con}(\graspemb, \wristemb, \negemb{g,1:K}) +\mathcal{L}_\text{con}(\graspemb, \placeemb, \negemb{g,1:K}) \\
+& \mathcal{L}_\text{con}(\placeemb, \wristemb, \negemb{p,1:K}) +\mathcal{L}_\text{con}(\placeemb, \graspemb, \negemb{p,1:K})
\\
+&\sum\mathcal{L}_{\text{reg}}(v) \text{ for } v \text{ in } [\graspemb, \placeemb, \wristemb,\negemb{g,1:K},\negemb{p,1:K} ]
\end{split}
\]
We train $\encbin$ and $\encee$ to minimize this loss function over the dataset of grasps using the Adam optimizer~\cite{kingma2014adam}.

\subsection{Choosing negative samples}
\label{sec:method_negative}
We propose two strategies for choosing the negative embeddings in the contrastive loss. The first is to use all other pixels in the spatial embedding as negatives, effectively performing a classification over the image. However, this pushes the embeddings at the selected pixels and its adjacent pixels away from each other even if they are on the same object. As an alternative, we sample distances from the selected pixel according to a Gamma distribution $\gamma(x; \alpha, \beta) = \beta^\alpha x^{\alpha-1}e^{-\beta x}/\Gamma(\alpha)$ and use pixels at those distances as negatives. $\Gamma(\alpha)$ is the gamma function equal to $(\alpha-1)!$ for integer $\alpha$. Parameters $\alpha=4$ and $\beta=w/8$, where $w$ is the image width, were set such that the number of negatives on the selected object are reduced. In our ablations in Section~\ref{sec:offline} we find that using both of these strategies together performs best.

\section{Using pixel-wise embeddings for goal-conditioned pick and place}
\label{sec:kitting}
We now present an algorithm for using the trained encoders $\encbin$ and $\encee$ to choose where to grasp and place without additional training. 
We address a goal conditioned pick and place task where the robot is given a goal image $\imgoal$ and a grasp bin image $\impart$ from which to pick objects and an empty place bin to place into $\imkit$. It must decide where to grasp and where to place to match the goal image. An example of this task is shown in Figure~\ref{fig:kitting}. We use the learned embedding space as a similarity metric over bin images
\[\text{sim}(\imkit, \imgoal) = \sum_x\sum_y {{\encbin(\imkit)^{(x,y)}} ^\top \encbin(\imgoal)^{(x,y)}}
\]
for all pixels $(x,y)$. A kit that has been successfully assembled to match a goal should have a very similar embedding map as the goal. %

\subsection{Grasping}
In order to assemble a kit that matches the goal, the robot should grasp an object out of the grasp bin that, when added to the place bin, improves $\text{sim}(\imkit, \imgoal)$. 
To find the best grasp position in the part tray $\gp$, we iterate over the pixels in each image to find the pixel leading to most similarity increase:
\begin{equation*}
\gp = \arg\max_{(x,y)}  [\max_{(i,j)}[  {\embpart{x,y}}^\top \embgoal{i,j}   - {\embkit{i,j}}^\top \embgoal{i,j} ] ].
\end{equation*}
which can be simplified as 
\begin{equation}
\label{eq:pick}
\gp = \arg\max_{(x,y)} [ \max_{(i,j)}[ ({\embpart{x,y}}-{\embkit{i,j}}) ^\top \embgoal{i,j} ]] 
\end{equation}
The intuition behind this objective is the idea that the pick and place process will replace the kit's embedding at position $(i,j)^*$ with the grasp bin's embedding at $\gp$.

\subsection{Placing}
After performing the grasp at $\gp$, we must decide what position to place at in the kit. One option is to place in the kit at pixel position $(i,j)^*$, which should be the optimal place to put the object that was picked. However, a more robust approach is to use the image from the wrist camera after the grasp $\imgripper$. As this image encodes the object as it was actually grasped, we can compare $\encee(\imgripper)$ to $\encbin(\imgoal)$ in order to decide where to place the grasped object. Using the wrist image allows the algorithm to correct for errors in the picking process. To choose the place position $\pp$, we search over the pixels of the kit to see where placing the object would most improve similarity to the goal:
\begin{equation}
\label{eq:place}
\pp = \arg\max_{(x,y)} {\embee}^\top (\embgoal{x,y} - \embkit{x,y})
\end{equation}
The whole pick and place algorithm can be iterated to move multiple objects into the kit.
\section{Experiments}
\begin{figure}[t]
    \begin{center}
	\adjustbox{max width=0.8\linewidth}{
        \begin{tabular}{|c c c c c c|}
        \hline
        & Banded & Blotchy & Braided & Dotted & Paisley\\
        Train&
            \includegraphics[width=0.1\linewidth]{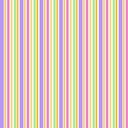} &
            \includegraphics[width=0.1\linewidth]{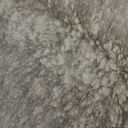} &
            \includegraphics[width=0.1\linewidth]{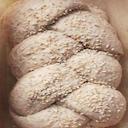} &
            \includegraphics[width=0.1\linewidth]{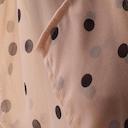} &
            \includegraphics[width=0.1\linewidth]{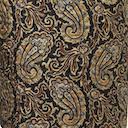} \\
        Val seen&
                \includegraphics[width=0.1\linewidth]{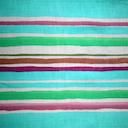} &
            \includegraphics[width=0.1\linewidth]{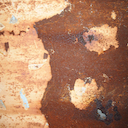} &
            \includegraphics[width=0.1\linewidth]{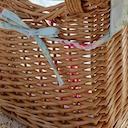} &
            \includegraphics[width=0.1\linewidth]{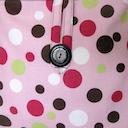} &
            \includegraphics[width=0.1\linewidth]{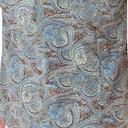} \\ \hline
        & Grid & Honeycombed & Marbled & Woven & Zigzagged\\
        {Val unseen} &
                \includegraphics[width=0.1\linewidth]{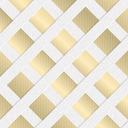} &
            \includegraphics[width=0.1\linewidth]{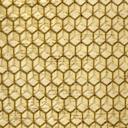} &
            \includegraphics[width=0.1\linewidth]{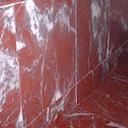} &
            \includegraphics[width=0.1\linewidth]{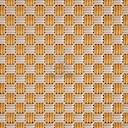} &
            \includegraphics[width=0.1\linewidth]{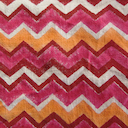} \\
            \hline
            
		\end{tabular}
	}
	\end{center}
	\caption{Example textures from the describable textures dataset that we use to texture objects in the grasping environment. The training set and validation seen set use textures from the same description classes, while validation unseen uses held out description classes.}
	\label{fig:textures}
\end{figure}

\begin{table*}[t]
    \centering

\begin{tabular}{|c|c|c|c|c|c|c|c|}
\hline
Num Train Textures & \gam & \fullimage & \grasplace&Train & Val Train & Val Seen  & Val Unseen  \\ 
\hline

10& \checkmark& & &23&22 & 20 & 21\\ \hline
10& \checkmark& &\checkmark& 20&19 & 19 & 18\\ \hline
10& &\checkmark& &51&29&31 &31 \\ \hline
10&\checkmark &\checkmark& &82&31&32  &34 \\ \hline
10& &\checkmark& \checkmark &23&26&30 &29 \\ \hline
10&\checkmark &\checkmark& \checkmark &\textbf{88}&\textbf{54}&\textbf{59} &\textbf{60} \\ \hline
\hline
50 &\checkmark & & &23& 24& 25& 24\\ \hline
50 & \checkmark& & \checkmark&25& 27 & 27  & 27\\ \hline
50& & \checkmark & & 70&60  & 59 & 59\\ \hline
50& &\checkmark&\checkmark& 45&29 & 29 & 28\\ \hline
50&\checkmark&\checkmark& &26& 34 & 31 & 32\\ \hline
50&\checkmark&\checkmark&\checkmark &\textbf{83}& \textbf{69}&\textbf{70} &\textbf{71}\\ \hline
\end{tabular}
    \caption{Ablations, all numbers are \% accuracy. We report the average object-level accuracy of our model in an offline evaluation. We find that best performance is obtained when \gam and \fullimage negative samplings are used together in addition to the \grasplace contrastive loss. We also find that generalization to new textures is no more difficult than generalization to held out grasp data using the same textures as in training set. Training on more textures leads to increased performance on all validation sets. }
    \label{tab:train_variants_results}
\end{table*}

For all datasets, we collect grasps in the simulated environment shown in Figure~\ref{fig:env} which is implemented in Unity. The environment does not simulate grasping dynamics: all surfaces of the objects are considered graspable\footnote{Note that the main focus of this work has been on learning visual embeddings and not on learning grasp dynamics, which we leave to future work.}. Data is collected by dropping 6 randomly chosen objects into the grasp bin and $\impick$ is saved from the camera above the grasp bin. A point on an object is randomly sampled as the grasp point $\gp$. The object is moved to the gripper by that point and $\imgripper$ is saved from the wrist-camera. Finally, a random point $\pp$ in the place bin is selected at which to place the object, the object is placed, and $\implace$ is saved from the camera above the place bin. The light color and brightness are randomized for each step in the data collection. The objects are sampled from a set of rectangular prisms textured with images from the describable textures dataset~\cite{cimpoi14describing}, with examples shown in Figure~\ref{fig:textures}. All training datasets contain 40000 grasps and places. The validation sets contain 500 grasps and places.

\subsection{Offline Evaluation}
\label{sec:offline}

We first evaluate our model in several offline settings. As we use textures from the describable textures dataset~\cite{cimpoi14describing}, we measure how our model generalizes to new textures both within and outside of the texture classes seen during training. As illustrated in Figure~\ref{fig:textures}, the training data and ``val seen" uses different textures form the same class, while the ``val unseen" uses textures from held out texture classes. The ``val train" setting evaluates the models on held out grasp data that uses the same textures as ``train".%

For these offline evaluations, we measure grasp and place accuracy by finding the pixel in $\embgrasp{}$ and $\embplace{}$, respectively, which has the largest dot product with the embedding of $\embee$ at the end-effector. If this pixel is on the correct object, the grasp or place is considered successful. We report the average of both grasp and place offline accuracy.

As discussed in Section~\ref{sec:method_negative}, we proposed two methods for sampling negatives: \fullimage, where all other pixels are used as negatives, and \gam  where negatives are sampled to a gamma distribution to avoid pushing apart the embeddings of nearby pixels. We ablate the necessity of contrasting the place embedding against the grasp embedding and vice versa (which we denote \grasplace), compared to only contrasting each against the wrist-cam image. The results in Table~\ref{tab:train_variants_results} show that using both negative sampling methods together along with the \grasplace contrastive loss leads to the best performance at both training and evaluation. 

Table~\ref{tab:train_variants_results} also compares training on just 10 different textures vs 50. We find that training on a greater variety of textures leads to better performance on all validation sets, even the set with the same textures as seen during training. Interestingly, while there is a drop in performance from 83\% accuracy on the training set of grasps to 69\% on the train-textures validation set, there is no drop from the train-textures validation to the unseen textures: all obtain about 70\% accuracy with a model trained on 50 different object textures. 

\subsection{Online evaluation}
We present preliminary results of using our self-supervised object embeddings in a goal-conditioned pick and place task. As shown in Figure~\ref{fig:kitting}, we apply our model to the goal-conditioned pick and place task described in Section~\ref{sec:kitting}. We freeze $\encbin$ and $\encee$ and use Equations~\ref{eq:pick} to determine where to grasp and place in order to make the kit match the goal. We find that the similarity metric learned by the embedding functions is highly sensitive to position within an object and can successfully select a correct object and place it into the right location. These preliminary results only show a single step of grasping/placing. Further work is needed to evaluate the multi-step case.

\section{Conclusion}
We presented a self-supervised approach to learning pixel-wise object-embeddings from random grasping data, and we developed a similarity metric based method for goal-conditioned kitting. We found that our embeddings generalize to unseen textures and unseen textures classes. Our work suggests that these embeddings can be learned from real robots as no part of training requires access to privileged information. We expect that this visual representation learning can be combined with self-supervised approaches for learning \emph{how} to grasp in order to extend this work to objects with more realistic grasp dynamics.

\bibliographystyle{plainnat}
\bibliography{references}

\end{document}